\documentclass[sigconf]{acmart}
\AtBeginDocument{%
  \providecommand\BibTeX{{%
    \normalfont B\kern-0.5em{\scshape i\kern-0.25em b}\kern-0.8em\TeX}}}

\setcopyright{acmcopyright}
\copyrightyear{2022}
\acmYear{2022}
\acmDOI{XXXXXXX.XXXXXXX}

\acmConference[Conference acronym CJO2022]{Make sure to enter the correct
  conference title from your rights confirmation emai}{August 15,
  2022}{Washington DC, USA}
\begin{document}

\title{A Large Scale Heterogeneous Treatment Effect Estimation Framework and Its Applications of Users' Journey at Snap}

\author{Li Shi}
\email{lshi@snapchat.com}
\affiliation{%
  \institution{Snap Inc.}
  \city{Santa Monica}
  \state{California}
  \country{USA}
}

\author{Jing Pan}
\email{jpan2@snapchat.com}
\affiliation{%
  \institution{Snap Inc.}
  \city{Santa Monica}
  \state{California}
  \country{USA}
}

\author{Paul Lo}
\email{plo@snapchat.com}
\affiliation{%
  \institution{Snap Inc.}
  \city{Santa Monica}
  \state{California}
  \country{USA}
}

\author{Jonathan Grotts}
\email{jgrotts@snapchat.com}
\affiliation{%
  \institution{Snap Inc.}
  \city{Santa Monica}
  \state{California}
  \country{USA}
}

\author{Crystal Pan}
\email{cpan@snapchat.com}
\affiliation{%
  \institution{Snap Inc.}
  \city{Santa Monica}
  \state{California}
  \country{USA}
}

\author{Ben Thompson}
\email{bpthomp@umich.edu}
\affiliation{%
  \institution{Snap Inc.}
  \city{Santa Monica}
  \state{California}
  \country{USA}
}

\author{Mattia Fumagalli}
\email{mfumagalli@snapchat.com}
\affiliation{%
  \institution{Snap Inc.}
  \city{Santa Monica}
  \state{California}
  \country{USA}
}

\author{Amit Adur}
\email{amit.adur@snapchat.com}
\affiliation{%
  \institution{Snap Inc.}
  \city{Santa Monica}
  \state{California}
  \country{USA}
}

\renewcommand{\shortauthors}{Li and Jing, et al.}

\begin{abstract}
    Heterogeneous treatment effects (HTE) and Conditional Average Treatment Effect (CATE) models break the often-used assumption that the treatment effect is uniform for each observation in an experiment. These models have been used with success in a variety of industrial applications. As advertising is the main revenue driver on most social media platforms, ensuring ads are seen by those who are most representative and responsive presents an essential way to improve the user journey. This paper describes a large scale industrial framework for estimating HTE, trained using experimental data covering hundreds of millions of users on Snapchat. The framework provides novel insights about customer interaction with the platform by combining experimental results across studies which reveals latent characteristics that were previously unmeasurable. We provide details on how to develop such a scalable framework, key findings on experiment selection criteria, base learner algorithms and incremental training that all significantly improved model performance. We also introduce two applications: user influenceability by ads and user sensitivity to ads to optimize users' journey on Snapchat. Initial online A/B test of applying the user influenceability scores in a targeting application has demonstrated an exceptional improvement on
    key business metrics that Snap tracks, by a factor of six on top of what is generally considered significant. 
\end{abstract}



\maketitle

\section{Introduction}
Heterogeneous treatment effects (HTE) and Conditional Average Treatment Effect (CATE) models have been used with success in a variety of industrial applications for content recommendation systems \cite{lada2019observational}, marketing \cite{zhao2019uplift}, and healthcare \cite{kravitz2004evidence}. A/B experimentation which fuels HTE estimation is the backbone of decision making for most technology companies. Given the velocity of policy decisions that are made across leading companies, thousands of experiments can be running simultaneously within a single application. Data from large scale experimentation platforms can be leveraged beyond high-level policy decisions by employing HTE methodology which can further be exploited by systematically combining data across multiple experiments. 

This paper describes a large scale industrial framework for estimating HTE, trained using experimental data covering hundreds of millions of users on Snapchat. We provide key insights on implementing such a scalable framework. The process to select experimental data for training is presented in Section~\ref{setup}. Subsequently, Section \ref{method} examines different uplift modeling methods and Section~\ref{sec:framework} introduces the details of the distributed/incremental training pipelines using TensorFlow Extended (TFX) platform~\cite{TFX2022}. 
This framework has been deployed in production and optimizes the incremental effects to serve more personalized and relevant ads during each user's journey with early results presented in Section \ref{results}. 

This framework has the potential to revolutionize the objective function of multiple technologies that are critical to the user journey at Snap. The large anticipated impact is based on transitioning from labels that are correlated with the desired outcome to labels that capture causal or incremental user behavior. Technology that can be influenced by this change includes but is not limited to ad ranking, lookalike and audience expansion, targeting, ad delivery, and content recommendation. 
In Section \ref{apps}, detailed applications are listed including users' influenceability and sensitivity to Snap’s advertising. The implementation of a scalable HTE framework can improve user journey on Snapchat in multiple dimensions.  

\section{Setup}  
\label{setup}
\subsection{Experiments Selection}
Relying on the randomization between treatment and control in experiments data to recover the counterfactual, we used the data collected from Snap's A/B test in direct-response advertising via the Intent-To-Treat (ITT) system to train the model. There are hundreds of experiments running daily and even more for historical experiments. Rather than blindly using all of our data, we carefully selected a subset that addresses several concerns, including concept drift (by selecting recency of the experiments) and low signal-to-noise ratio (by selecting the control group size and relative lift). As an example, experiments that ended more than 180 days ago or have less than 10k users in the control group were excluded. These parameters were carefully tuned to select the best performing model based on metrics described in the subsequent section. Even with these restrictions, our training data covered billions of dollar in ad spend, hundreds of millions of users, and around 1,000 features for each observation. The model was evaluated on a large amount of studies to ensure the robustness and conciseness of the model.

To tune the model's performance, we exhaustively grid searched the criteria of experiment candidates during model training. We also evaluated on a large amount studies (1.2 billion observations) to ensure the robustness and conciseness of the model. Our analysis shows that by selecting more recent experiments with larger relative lift, we were able to obtain better model performance, details are summarized in Section~\ref{subsec:experiment_selection}.

\subsection{Metrics}
When learning uplift models in a real-world setting, we don’t have the ground truth due to the fact that we cannot observe a user’s incrementality directly (both the control and the treatment outcomes for each observation). Hence, we focused on the internal validation methods under the assumption of unconfoundedness of potential outcomes and the treatment status conditioned on the feature set (around 1,000 features) available to us.

First, we validated the model by comparing the estimates with different approaches and checking the consistency of estimates across different levels, user groups, and time periods for model's robustness and stability.
\begin{itemize}
\item {\verb|Robustness|}: We compared the estimates from different underlying ML algorithms (e.g., linear regression, Deep Neural Network, Deep and Cross Network, Wide and Deep etc.). In addition, we split the data within a cohort and compared the results from out-of-sample estimates and in-sample estimates. 
\item {\verb|Stability|}: Treatment effect may vary from cohort to cohort and time to time but should not be too volatile. We evaluate the consistency across cohorts and time periods. 
\end{itemize}

Then, we evaluated the model by comparing the metrics of  Uplift (lift/gain/gini etc.) Curve \cite{gutierrez2017causal, rzepakowski2012decision, soltys2015ensemble, zhao2017uplift} and the Area Under the Uplift Curve (AUUC). Similar to Area Under the Curve (AUC), AUUC measures the cumulative lift (gain) after sorting  the predicted Individual Treatment Effect (ITE) from high to low. A well performing model (with a high AUUC) scores higher gain in those individuals for which the ITE is high (beneficial) compared to ones for which the ITE is low (neutral or even detrimental). We used these metrics to compare different models as well as random targeting to select the best performing one.

\section{Methodology}
\label{method}
\subsection{Uplift Models}
Various uplift models have been developed and implemented \cite{chen2020causalml} recently including but not limit to deep models (DRNet \cite{hassanpour2019learning}, SITE \cite{yao2018representation}, CEVAE \cite{louizos2017causal}, DragonNet \cite{shi2019adapting} etc.) and meta-learners (S \cite{kunzel2019metalearners}, T \cite{hansotia2002incremental, kunzel2019metalearners}, X \cite{kunzel2019metalearners} and R-learner \cite{nie2021quasi}, and DR-learner \cite{bang2005doubly} etc.). We are using the meta-learners T-learner/two model framework for three main reasons. First, the experimental data provides the uncounfededness that is intrinsically leveraged by the two-model framework. Second, it's intuitively easy to understand and explain to stakeholders or downstream customers how the ITE has been calculated. Finally, it also provides a flexible set-up to leverage all kinds of ML models as the base learners for optimizing model performance. Instead of directly using the deep learning models or traditional linear or gradient based decision tree models as base learners, we experimented with different deep learning models (e.g. Deep and Cross Network \cite{wang2017deep}, Wide \& deep Network \cite{cheng2016wide}) as the base learners. Through this framework, we were able to quickly iterate on base learners to find the model that gave the best performance, detailed results shown in Section~\ref{subsec:base_learners}.

\begin{figure}[h]
  \centering
  \includegraphics[width=\linewidth]{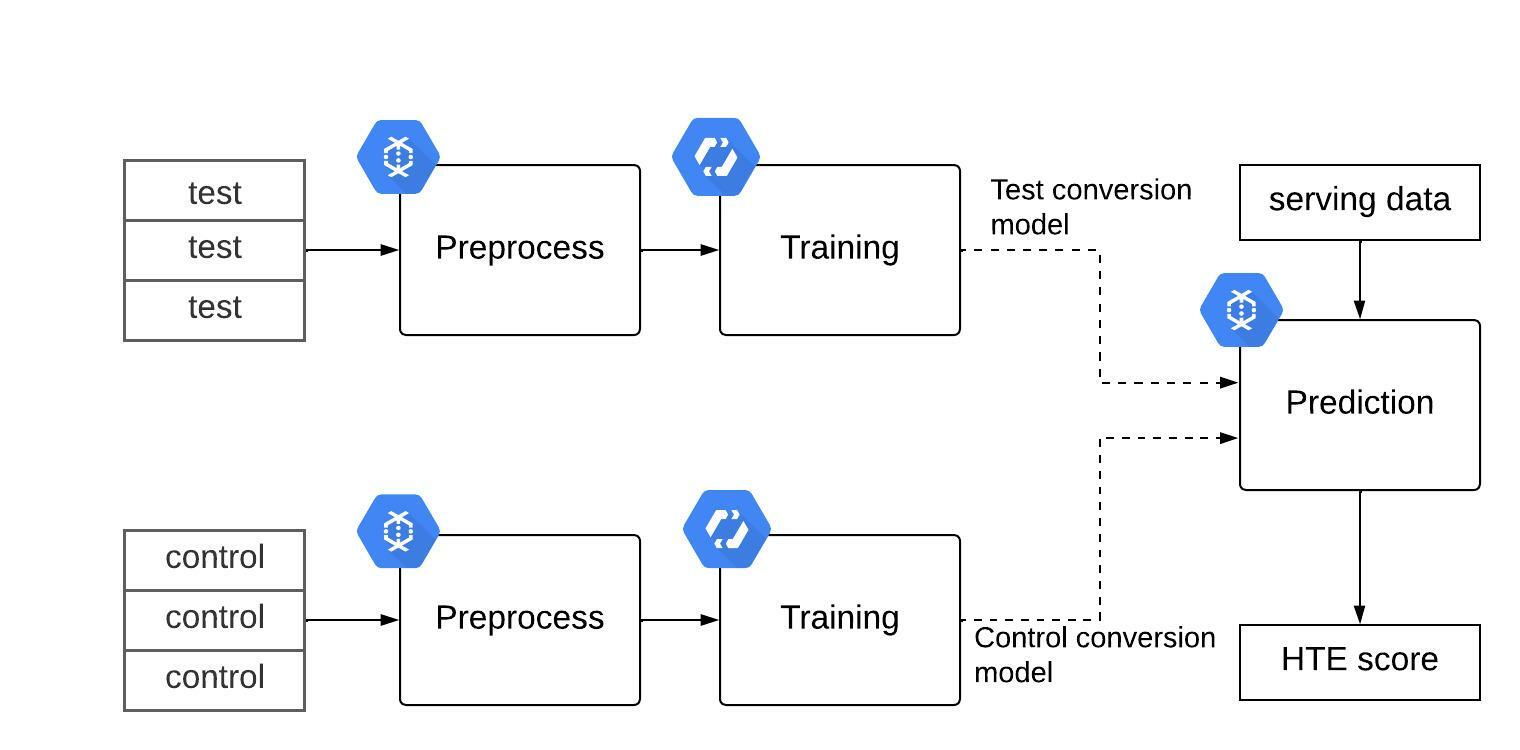}
  \caption{Overall workflow for a single training and serving}
  \label{fig:pite_framework_workflow}
\end{figure}

\section{Framework}
\label{sec:framework}
\subsection{Distributed System}
A general pain point for uplift models to be used in production system is the scalability. To be able to train the model with a large amount of experiments and deploy model serving for millions of users, the framework utilized TensorFlow Extended (TFX) platform \cite{TFX2022} for distributed data processing and feature engineering and Google AI platform for distributed model training and serving. TFX is a Google-production-scale machine learning platform based on TensorFlow. It provides a configuration framework and shared libraries to integrate common components needed to define, launch, and monitor the machine learning system. Our pipelines used Google AI platform’s training and serving service which scaled the framework for over hundreds of models for various applications. As shown in Figure~\ref{fig:pite_framework_workflow}, a typical training and serving workflow includes the preprocessing and training T-learner (test and control ML models) and the prediction step in which both test and control model are invoked. 

\begin{figure}[h]
  \centering
  \includegraphics[width=\linewidth]{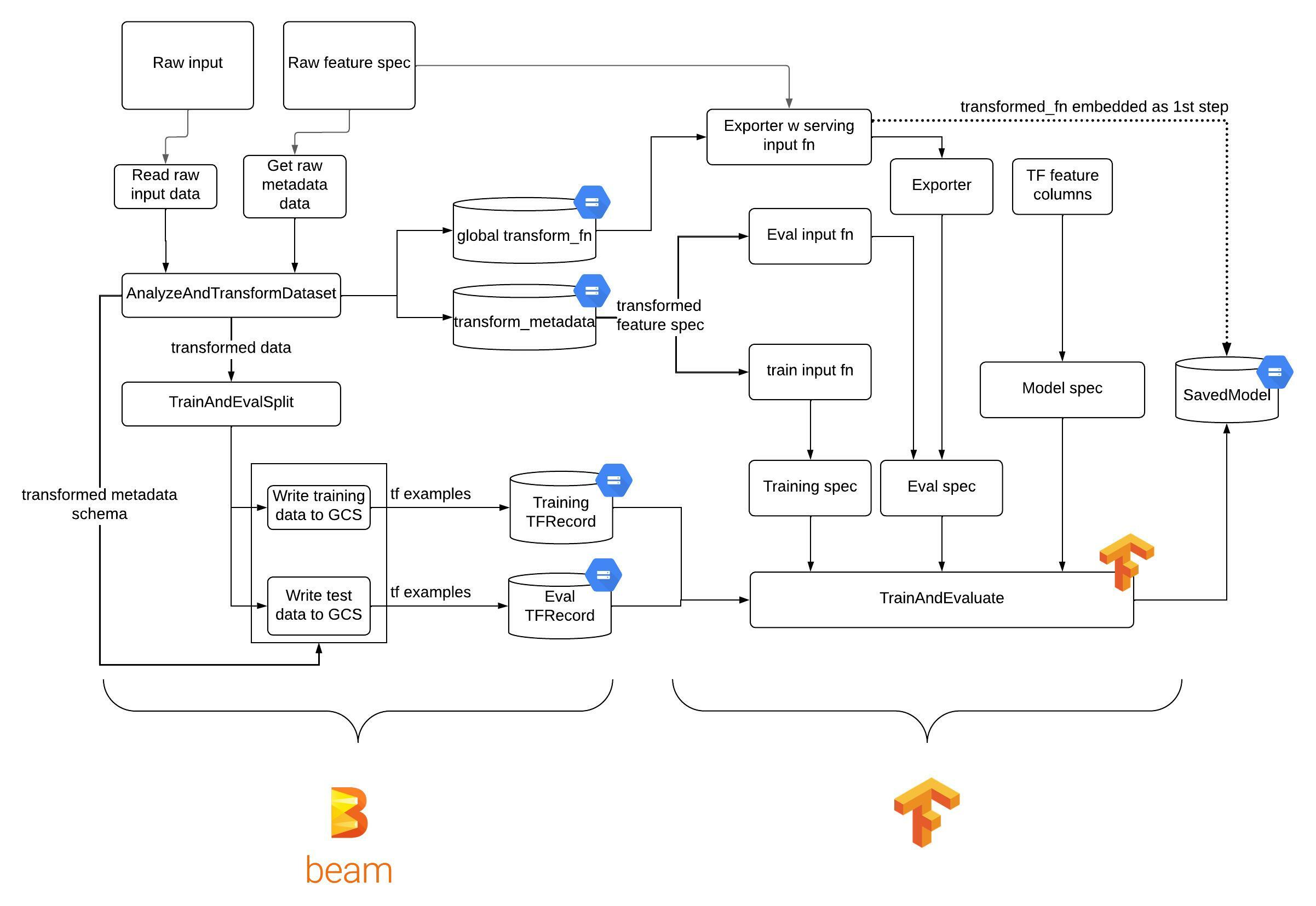}
  \caption{Detailed workflow for preprocessing and training}
  \label{fig:pite_preprocess_train}
\end{figure}

Figure~\ref{fig:pite_preprocess_train} shows the detailed workflow of preprocessing and training. In the preprocessing step, the raw input data, containing features and labels is read and transformed into TensorFlow (TF) Record which are training-ready. The transformed TFRecord are further split into training and validation datasets and written into Google Cloud Console (GCS) storage bucket. Besides the training and validation data, the transformation functions and transformed feature specifications are also written to GCS storage buckets. The transformation functions are applied to input data at the serving step to ensure an identical transformation is applied to training and prediction datasets.

\begin{figure}[h]
  \centering
  \includegraphics[width=\linewidth]{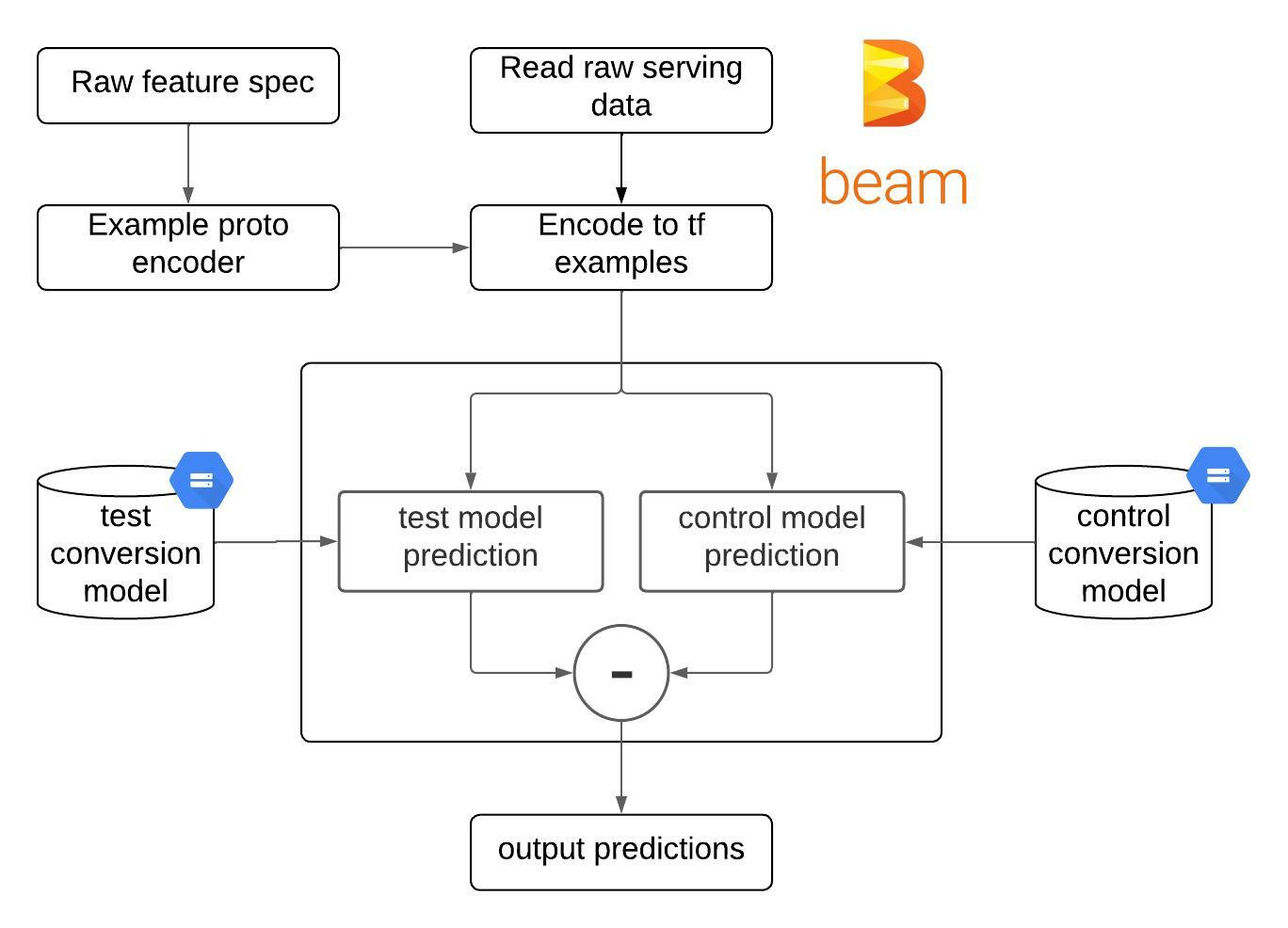}
  \caption{Detailed workflow for serving}
  \label{fig:pite_prediction}
\end{figure}

Figure~\ref{fig:pite_prediction} shows the details of the prediction step in which the serving data is read and encoded into TFRecord using raw feature specifications. The trained T-learner (test and control outcome models) are applied on each input. 

By having this flexible and scalable framework, it enables the faster model developing and deploying cycle and also provides the ability and possibility for using various algorithms and extending to applications very easily. Note that results presented in Section~\ref{results} and applications introduced in Section~\ref{apps} all used this framework.

\subsection{Incremental Training}
Incremental and online machine learning is a pertinent contemporary topic especially in the context of learning from real-time data streams. User behavior, advertiser spending, ad products, and a variety of other platform characteristics can change over time. A flexible training framework is necessary to capture these various moving pieces. This is in contrast to traditional assumptions of complete data availability. Classical batch machine learning approaches do not continuously integrate new information into already constructed models, but instead regularly reconstruct new models from scratch. This is not only very time consuming but also leads to potentially outdated models.

We propose an incremental training approach for learning and adapting to the new individual customer behaviors and environments in the system without forgetting previously learned information. In each week’s training pipeline, we incrementally train the existing models by incorporating the most recently completed experimental data. Feature and label monitoring is layered on to provide transparency into possible feature drift and outcome distribution changes.
A comparison between incremental training and regular weekly training is included in Section~\ref{subsec:eval_inc_training}.

\section{Experiments and Results} 
\label{results}

\begin{table}
  \caption{The average evaluation AUUC Score for experiment selection}
  \label{tab:xp_auuc}
  \begin{tabular}{ p{1cm}|p{1cm}|p{1cm}|p{1cm}|p{1.5cm}|p{1.5cm}|}
    \toprule
    Version & Recency (days)&Relative Lift&Control Size& $\#$ of Evaluation Study & AUUC\\
    \midrule
    V1 & 90 & $-$ & $-$ & 160 & 0.48 $\pm$ 0.06\\
    V2 & 90 & $>\delta$ & $>10K$ & 160 & 0.53 $\pm$ 0.06\\
    V3 & 180 & $>\delta$ & $>10K$ & 160 & 0.43 $\pm$ 0.05\\
    V4 & 90 & $>2\delta$ & $>10K$ & 160 & 0.54 $\pm$ 0.06\\
    V5 & 90 & $>3\delta$ & $>10K$ & 160 & \textbf{0.56 $\pm$ 0.06}\\
  \bottomrule
\end{tabular}
\end{table}

\begin{figure}[h]
  \centering
  \includegraphics[width=\linewidth]{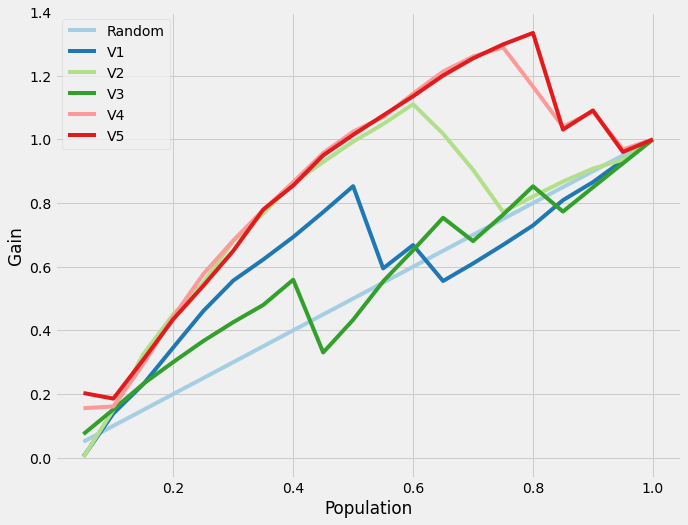}
  \caption{Uplift Curve for Different Experiment Selection Methods}
  \label{fig:xp_selection_uplift_curve}
\end{figure}

\subsection{Experiment Selection}
\label{subsec:experiment_selection}
Experiments selected to train the HTE models has a critical impact on the performance of the models. 
To achieve the best performance, we examined 5 different experiment selection methods which use different criteria on look-back time period, experiment treatment effect and experiment size. 
We performed out-of-sample evaluation on these methods using 160 real experiments ran on Snap. 
Table \ref{tab:xp_auuc} summarized the average test AUUC scores with the 95\% confidence intervals.
As shown in table 1, the method V5, which selects the most recent 90 days experiments with more than three delta of relative lift, larger than 10k control user, yields the best performance. 
To illustrate the different abilities of capturing user-level heterogeneous treatment effect of models using these 5 different selection methods, as an example, Figure~\ref{fig:xp_selection_uplift_curve} shows the uplift curves of different methods on an evaluation experiment.

To summarize, through examining different ways to select experiments as training data, we found that tuning the recency, treatment effect and sample size of experiments generally result a better model performance.

\subsection{Base Learners Comparison}
\label{subsec:base_learners}
The base learner used in the HTE models is another crucial factor on the performance of the model. 
To choose the best base learner, we evaluated three different algorithms: Deep Neural Network, Wide and Deep Network, and Deep and Cross Network, using 35 experiments. 

\begin{figure}[h]
  \centering
  \includegraphics[width=\linewidth]{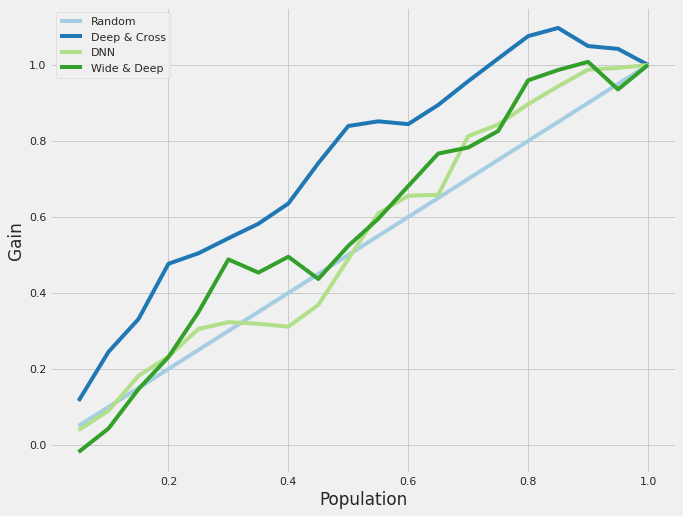}
  \caption{Uplift Curve for Different Base Learners}
  \label{fig:base_learner_uplift_curve}
\end{figure}

Table~\ref{tab:base_auuc} shows the average AUUC scores with the 95\% confidence intervals of the HTE models using the three different base learners. As shown in the table, the Deep and Cross Network demonstrates a superior performance over the other two base learner. 
To better visualize this, Figure~\ref{fig:base_learner_uplift_curve} shows the uplift curves of an evaluation experiment generated by the thre base learners.
As a result, the Deep and Cross Network is used as base learner in the production models.

\begin{table}
  \caption{The average evaluation AUUC Score for base learners}
  \label{tab:base_auuc}
  \begin{tabular}{c|c|c}
    \toprule
    Model Type & $\#$ of Evaluation Study & AUUC\\
    \midrule
    Deep and Cross Network & 35 & \textbf{0.59 $\pm$ 0.11}\\
    Deep Neural Network & 35 & 0.51 $\pm$ 0.10\\
    Wide and Deep & 35 & 0.50 $\pm$ 0.11 \\
  \bottomrule
\end{tabular}
\end{table}

\subsection{Incremental Training}

\begin{figure}[h]
  \centering
  \includegraphics[width=\linewidth]{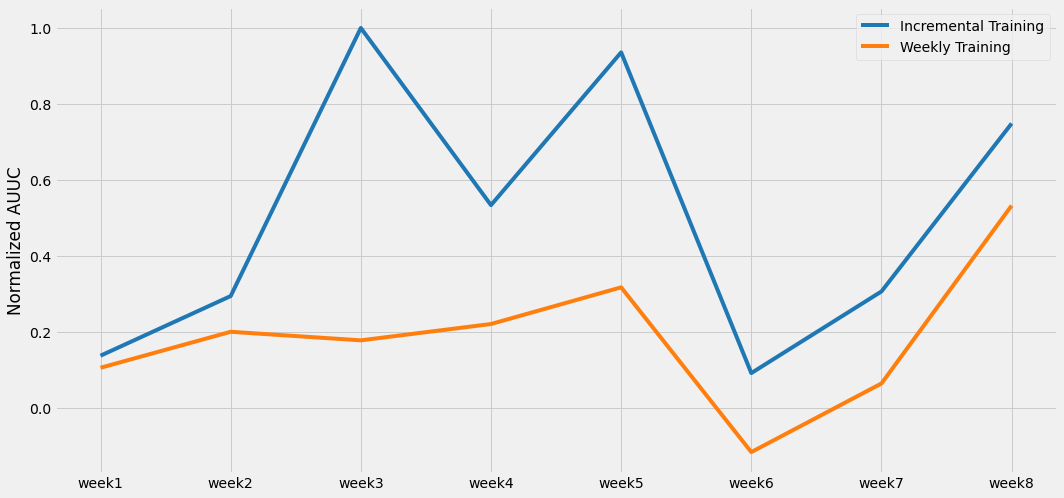}
  \caption{Week over Week AUUC: Incremental Training vs. Weekly Deep and Cross Network Training}
  \label{fig:incremental_training_wow}
\end{figure}

Models used in production are re-trained weekly.We compared the regular weekly training with the incremental training over multiple weeks on 18 in production models which both used the same experiment selection method and base learners (Deep and Cross Network). At the beginning of the training, two approaches yielded similar performance, but after two weeks online incremental trained models started outperforming the weekly DCN models. Figure \ref{fig:incremental_training_wow} shows consistent better week over week performance with even more than 100\% improvement for the average AUUC scores .We observed that instead of fine tuning the model on specific datasets, incremental training utilized both historical and recent datasets for a superior performance.

\label{subsec:eval_inc_training}

\section{Applications} 
\label{apps}
\subsection{Overview}
Snapchat is a platform that serves hundreds of millions of users, with an aim to deliver meaningful impact to the users and advertisers, we establish a holistic HTE estimation framework which covers different aspects in applications and touch points of the User Journey, from the incrementality of conversion metrics to potential drop in engagement metrics when ads load and relevancy might deteriorate overtime.  

\subsection{User Influenceability by Ads}
User influenceability is a metric to measure the incrementality a given user will make a conversion because of seeing ads on Snapchat. This is an essential metric as our goal is to serve more personalized and relevant ads to users and at the same time prove the ads value to advertiser as well. Meanwhile, the focus on measuring the incrementality of seeing an ad and the volume of such advertisers is rapidly increasing.

\begin{figure}[h]
  \centering
  \includegraphics[width=\linewidth]{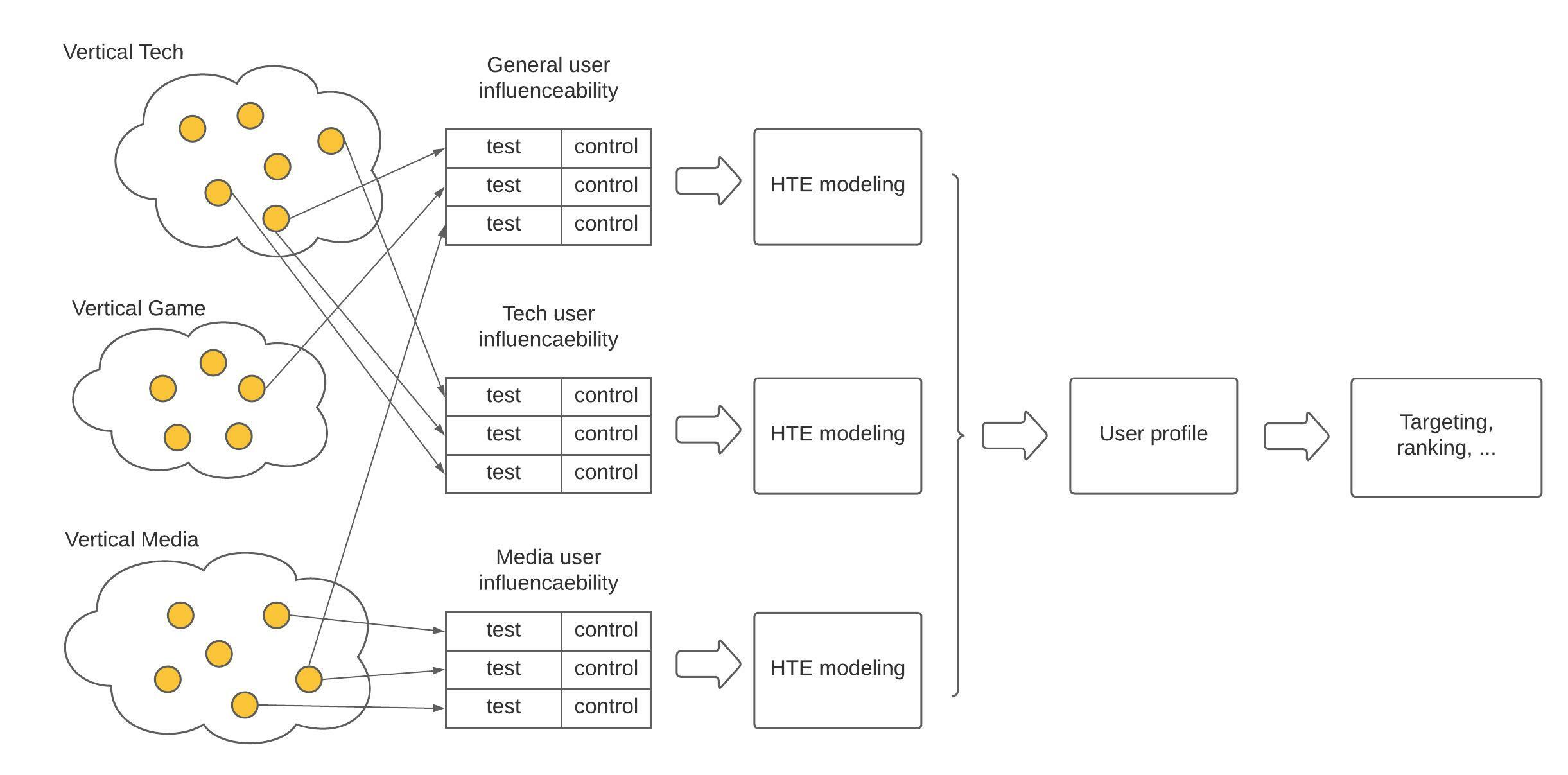}
  \caption{User Influenceability model structure}
  \label{fig:user_influenceability_scope}
\end{figure}

User influenceability can be measured at different scope, from general purpose to advertiser specific, as shown in Figure~\ref{fig:user_influenceability_scope}. There are applications for both in our system to improve user journey with very promising results. 
We deployed one general model and 10 vertical level models which generates HTE estimation on hundreds of millions of users weekly. The scores are integrated with our targeting tools, demonstrating an exceptional improvement on key business metrics Snap tracks internally, by a factor of six on top of what is generally considered significant. Meanwhile, the dvertiser-specific ITE score can be used by specific advertisers through targeted audiences therefore there will be more relevant and personalized content/ads delivered during the user journey.

\subsection{User Sensitivity to Ads}
Besides user influenceability by ads, the impact of ads on user engagement is another important factor in ads delivery system, acting as a metric that retains health user engagements and long-term revenue. To measure the negative impact of ads on user engagement, at Snap, we ran no ads holdout A/B experiments. Users are controlled from seeing specific types of ads to measure the impact of different user engagement metrics.

\begin{figure}[h]
  \centering
  \includegraphics[width=\linewidth]{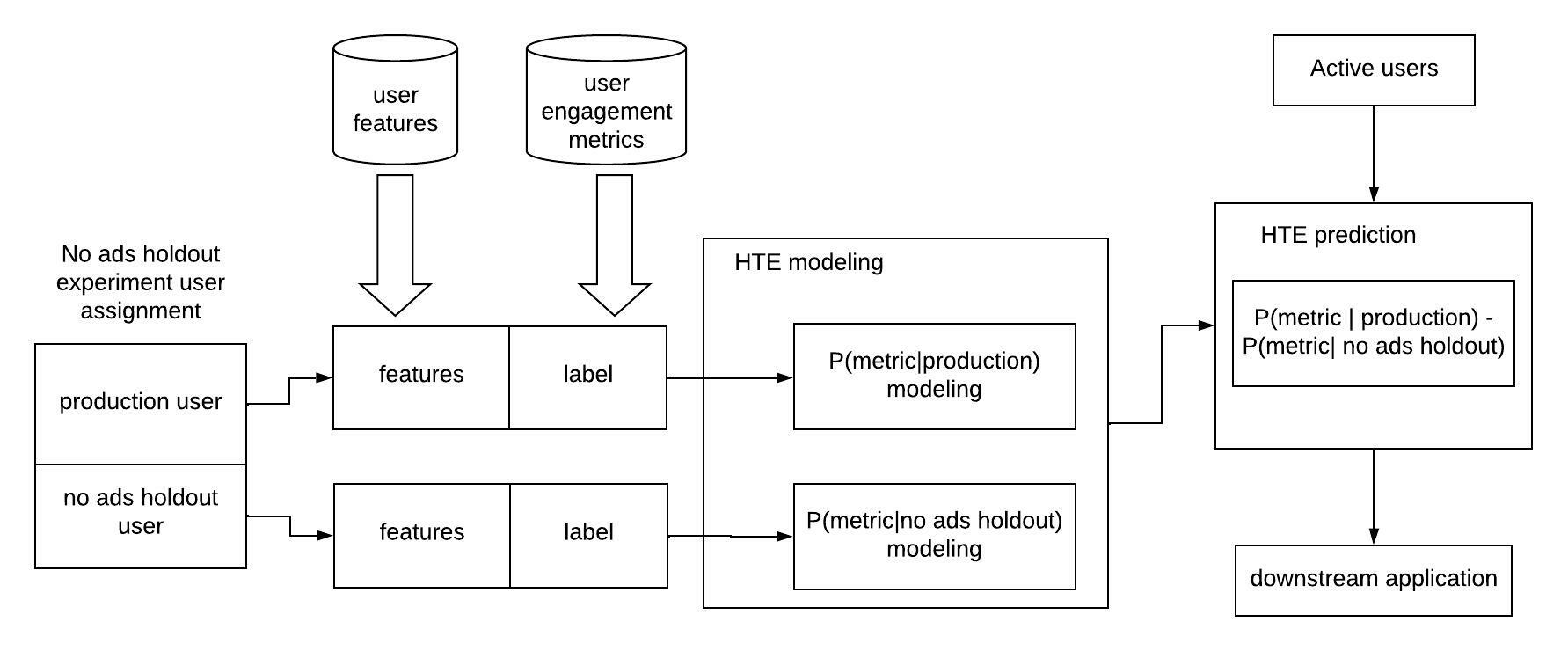}
  \caption{User Sensitivity model structure}
  \label{fig:pite_rm_high_level}
\end{figure}

Applying HTE models on those experiments allows us to estimate the reduction on user engagement caused by a particular ad product at user level. We denoted this user-level metric as User Sensitivity to ads. Figure~\ref{fig:pite_rm_high_level} shows the model structure of the User Sensitivity.

With a combination of a set of selected ad products and a set of key user engagement metrics, we deployed 18 user sensitivity models on hundreds of millions of users weekly. We observed 20\% average AUUC improvement on those models comparing with random consistently. 

Estimating the incremental user engagement metrics enable us to make our personalized ads display more optimized towards user experience, therefore increasing the future inventory though higher user engagement and generating more revenue in long-term. For example, when deciding the candidate ads to show for an ad slot of individual user, user sensitivity scores can provide a finer-granular adjustments on the value of each candidate ad. 

\section{Conclusion and Next Steps} 
This paper demonstrates that a large scale framework for estimating heterogeneous treatment effects can improve the customer journey at a social media company. The details on implementation are provided to measure causally motivated metrics for downstream applications. The system is able to measure HTE on hundreds of millions of users. The models, trained on experimental data from multiple advertiser campaigns, has been deployed successfully on different business applications in a highly extensible manner. The focus for the initial application was improving users experience on the platform by presenting them with relevant advertisement based on historical user behavior. For user influenceability, we effectively deployed a 'wedding cake' set of models, that cover user response to all ads, to ads in a particular industry, and to ads for particular (selected) client.

We optimized the approach by testing different experiment selections and base learners in the uplift modelling, including a weekly incremental training method, and we have summarized insights to drive better performance. The integration of framework in real-world ads targeting, and its increase in performance from the HTE features have demonstrated tremendous value and impact.

Future improvements to the framework include continued experimentation with the incremental training approach and different uplift modeling algorithms. Additional attention is needed to address recent privacy restrictions by mobile platforms which has disrupted the attribution process. These disruptions have made it impossible to track user level events in a deterministic way which largely impacting the quantity and qualify of the model training data. Training high-quality uplift models in this environment with Snap's privacy-centric approaches will be a focus in our research efforts.

\begin{acks}

\end{acks}

\bibliographystyle{ACM-Reference-Format}
\bibliography{sample-base}

\end{document}